\documentclass[10pt,twocolumn,letterpaper]{article}

\usepackage{cvpr}
\usepackage{times}
\usepackage{epsfig}
\usepackage{graphicx}
\usepackage{enumitem}
\usepackage{amsmath}
\usepackage{amssymb}
\usepackage[]{booktabs}
\usepackage{multirow}
\usepackage{tabularx}
\usepackage[table,xcdraw]{xcolor}
\usepackage{cite}
\usepackage[pagebackref=true,breaklinks=true,letterpaper=true,colorlinks,bookmarks=false]{hyperref}

 \cvprfinalcopy 

\def\cvprPaperID{****} 

\ifcvprfinal\fi
\begin{document}

\title{Convolutional Social Pooling for Vehicle Trajectory Prediction}

\author{Nachiket Deo \quad Mohan M. Trivedi\\
University of California, San Diego \\
La Jolla, 92093\\
{\tt\small ndeo@ucsd.edu \quad mtrivedi@ucsd.edu}
}

\maketitle

\begin{abstract}
Forecasting the motion of surrounding vehicles is a critical ability for an autonomous vehicle deployed in complex traffic. Motion of all vehicles in a scene is governed by the traffic context, i.e., the motion and relative spatial configuration of neighboring vehicles. In this paper we propose an LSTM encoder-decoder model that uses convolutional social pooling as an improvement to social pooling layers for robustly learning interdependencies in vehicle motion. Additionally, our model outputs a multi-modal predictive distribution over future trajectories based on maneuver classes. We evaluate our model using the publicly available NGSIM US-101 and I-80 datasets. Our results show improvement over the state of the art in terms of RMS values of prediction error and negative log-likelihoods of true future trajectories under the model's predictive distribution. We also present a qualitative analysis of the model's predicted distributions for various traffic scenarios. 
\end{abstract}

\section{Introduction}

In order to safely and efficiently navigate through complex traffic comprised by human drivers, an autonomous vehicle needs to have the ability to take initiative, such as deciding when to change lanes, overtake another vehicle, or slowing down to allow other vehicles to merge. This requires the autonomous vehicle to have some ability to reason about the future motion of surrounding vehicles. This can be seen in existing tactical path planning algorithms \cite{ulbrich,nilsson,dpdm}, which depend on reliable estimation of future trajectories of surrounding vehicles.
\begin{figure}[t]
\centering
\includegraphics[width=\columnwidth]{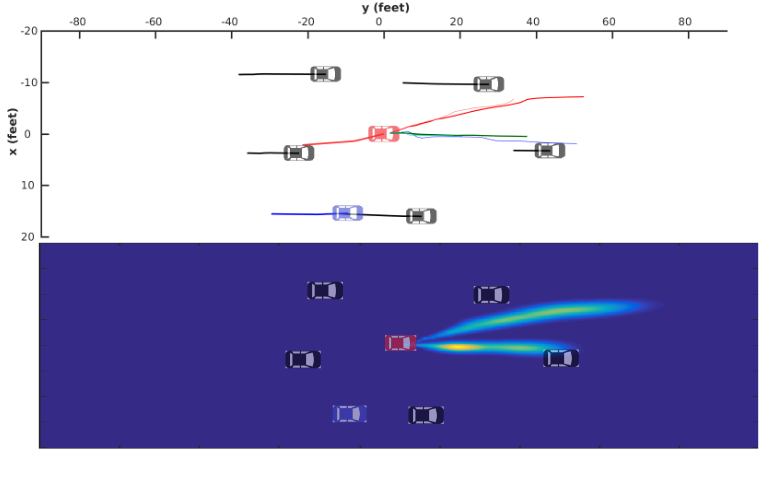}
\caption{Imagine the blue vehicle is an autonomous vehicle in the traffic scenario shown. Our proposed model allows it to make multi-modal predictions of future motion of it's surrounding vehicles, along with prediction uncertainty shown here for the red vehicle }
\label{figurelabel}
\end{figure}

Prediction of future motion of surrounding vehicles is a challenging problem due to the high number of latent variables involved, such as, the end goals of all drivers in the scene and variability in driving style across different drivers. Vehicle trajectories tend to be highly non-linear over longer time horizons due to decisions made by the driver. Additionally, driver behavior tends to be inherently multi-modal, where a driver could make one of many decisions under the same traffic circumstances. Finally, interaction between vehicles tends to affect their motion. The large number of possible configurations of all vehicles in a scene can make this difficult to model.

In spite of these challenges, there is structure to vehicle motion that can be exploited:
\begin{itemize}[itemsep=1pt,parsep=0.3pt]
\item Vehicle motion can be binned into \textit{maneuvers}, which can account for the multi-modal nature of future motion. For example, a vehicle approaching its leading vehicle at a much faster speed would either brake and slow down, or change lane to overtake.    
\item There is a well defined lane structure and a direction of motion on freeways. This can be exploited to model interaction between vehicles   
\end{itemize}

Following the success of long-short term memory (LSTM) networks in modeling non-linear temporal dependencies in sequence learning and generation tasks \cite{soclstm,graves,cho}, we propose an LSTM encoder-decoder based model for vehicle motion prediction for the case of freeway traffic. In particular, our model can be characterized by:
\begin{enumerate}[itemsep=1pt,parsep=0.3pt]
\item \textbf{Convolutional social pooling:} We propose a novel social pooling layer as an alternative to that proposed in \cite{soclstm}. We apply convolutional and max-pooling layers instead of a fully connected layer to \textit{social-tensors} of LSTM states that encode the past motion of neighboring vehicles. 
\item \textbf{Maneuver based decoder:} Our LSTM decoder generates the probability distribution over future motion for six maneuver classes and assigns a probability to each maneuver class. This accounts for the multi-modal nature of vehicle motion.   
\end{enumerate}

\section{Related Research}
An extensive survey on vehicle motion prediction models has been presented by Lefevre \textit{et al.} \cite{surv}, where the the models are categorized into physics based, maneuver based and interaction aware models. Close to our approach are the maneuver based models and interaction aware models.\\

\noindent\textbf{Maneuver based models:}
Classification of vehicle motion into semantically interpretable maneuver classes has been extensively addressed in both advanced driver assistance systems as well as naturalistic drive studies. Of particular interest are works that use the recognized maneuvers to make better predictions of future trajectories \cite{houenou,schreier,tran,laug,gmrlat,self}. These approaches usually involve a maneuver recognition module for classifying maneuvers and maneuver specific trajectory prediction modules. Maneuver recognition modules are typically classifiers that use past positions and motion states of the vehicles and context cues as features. Heuristic based classifiers \cite{houenou}, Bayesian networks \cite{schreier}, hidden Markov models \cite{laug,self}, random forest classifiers \cite{gmrlat} and recurrent neural networks have been used for maneuver recognition. Trajectory prediction modules output the future locations of the vehicle given its maneuver class. Polynomial fitting \cite{houenou}, maneuver specific motion models \cite{schreier}, Gaussian processes \cite{tran,laug}, and Gaussian mixture models \cite{self} have been used for trajectory prediction.\\     

\noindent\textbf{Interaction aware models:}
Interaction aware models for motion prediction take into account the effect of inter-vehicle interaction on the motion of vehicles. Two different approaches can be found for incorporating inter-vehicle interaction. The first set of approaches \cite{self, bahram} use hand crafted cost functions based on the relative configuration of vehicles
and make optimal predictions of future motion with respect to these cost functions.  Cost function based approaches do not depend on training data and
can generalize to new traffic configurations. However, they can be limited by how well the hand-crafted cost function is designed. The second approach to incorporate inter-vehicle interaction is to implicitly learn it from trajectory data of real traffic. However, due to the large variation in traffic configurations, this approach requires a large dataset for generalization. This approach has been used in prior works for the case of two vehicles approaching an intersection \cite{vi1}, and lateral motion prediction on highways \cite{gmrlat}. We use the data-driven approach for inter-vehicle interaction in this paper, since it it not limited by the design of a hand-crafted cost function, and also due to the availability of large datasets of real freeway traffic \cite{ngsim1,ngsim2}.\\

\noindent\textbf{Recurrent networks for motion prediction:}
Since motion prediction can be viewed as a sequence classification or sequence generation task, a number of recurrent neural network (RNN) based approaches have been proposed in recent times for maneuver
classification and trajectory prediction. Khosroshahi \textit{et al.} \cite{aida} and Phillips \textit{et al.} \cite{phil} use LSTMs to classify vehicle maneuvers at intersections. Kim \textit{et al.} \cite{kim} propose an LSTM that predicts the location of vehicles in an occupancy grid at intervals of 0.5 s, 1 s and 2 s into the future. Contrary to this approach, our model outputs a continuous, multi-modal probability distribution of future locations of the vehicles up to a prediction horizon of 5 s. Lee \textit{at al.} \cite{desire} propose a model that combines conditional variational auto-encoders (CVAE) with RNN encoder-decoders for trajectory prediction. While this allows for multi-modal predictions by sampling the CVAE, the model can only provide samples from the predictive distribution rather than an estimate of the distribution itself. In their seminal work, Alahi \textit{et al.} \cite{soclstm} propose \textit{social LSTMs}, which jointly model and predict the motion of pedestrians in dense crowds through the use of a social pooling layer. We improve upon this approach by using convolutional social pooling. We also incorporate the lane structure of freeways into our social pooling layer. Finally, Kuefler \textit{et al.} \cite{gail} use a gated recurrent unit (GRU) based policy using the behavior cloning and generative adversarial imitation learning paradigms to generate the acceleration and yaw-rate values of a bicycle model of vehicle motion. We compare our trajectory prediction results with those reported in \cite{gail}.

\section{Problem Formulation}

We formulate motion prediction as estimating the probability distribution of the future positions of a vehicle conditioned on its track history and the track histories of vehicles around it, at each time instant $t$.
\subsection{Frame of reference}
\label{sec:for}
\begin{figure}[t]
\centering
\includegraphics[width=\columnwidth]{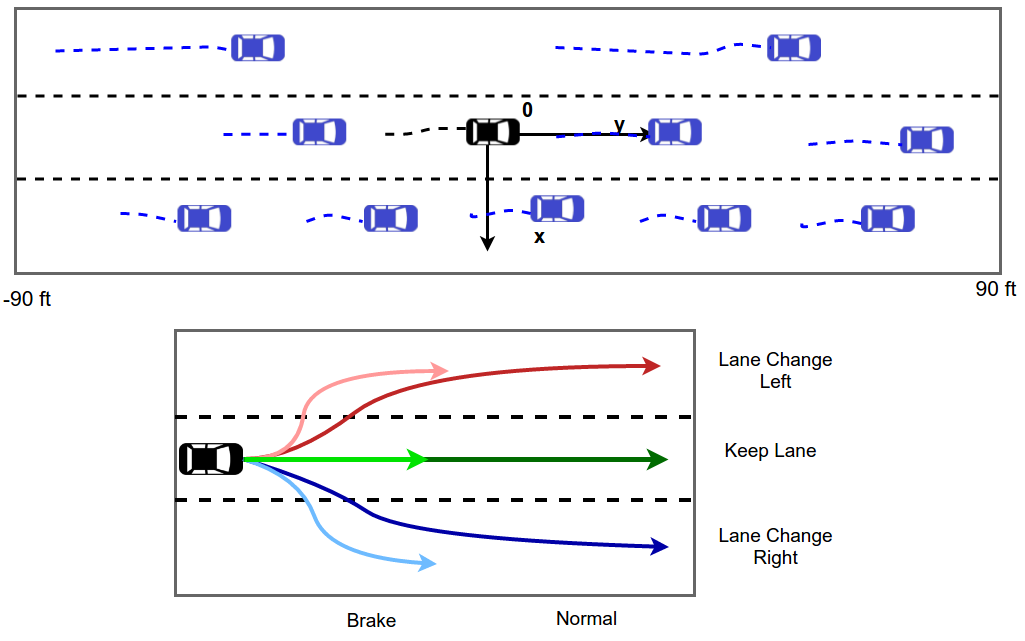}
\caption{\textbf{Top:} The co-ordinate system used for trajectory prediction. The vehicle being predicted is shown in black, neighboring vehicles considered are shown in blue. \textbf{Bottom:} Lateral and longitudinal maneuver classes}
\label{fig:form}
\end{figure}

 We use a stationary frame of reference, with the origin fixed at the vehicle being predicted at time $t$ as shown in Fig. \ref{fig:form}. The \textit{y-axis} points in the direction of motion of the freeway, and the \textit{x-axis} is the direction perpendicular to it. This  makes our model independent of how the vehicle tracks were obtained, and in particular, can be applied to the case of on-board sensors on an autonomous vehicle. This also makes the model independent of the curvature of the road, and can be applied anywhere on a freeway as long as an on-board lane estimation algorithm is available. 
\subsection{Inputs and outputs}
The input to our model are track histories
\begin{equation}
\mathbf{X} = [\mathbf{x}^{(t-t_{h})},...,\mathbf{x}^{(t-1)}, \mathbf{x}^{(t)}]
\end{equation}

\noindent where,
\begin{equation}
\mathbf{x}^{(t)} = [x_{0}^{(t)},y_{0}^{(t)},x_{1}^{(t)},y_{1}^{(t)},...,x_{n}^{(t)},y_{n}^{(t)}]
\end{equation}

\noindent are the $x$ and $y$ co-ordinates at time $t$ of the vehicle being predicted and all vehicles within $\pm  90$ feet in the longitudinal direction and within the two adjacent lanes of the vehicle being predicted, as shown in Fig. \ref{fig:form}.

The output of the model is a probability distribution over
\begin{equation}
\mathbf{Y} = [\mathbf{y}^{(t+1)},..., \mathbf{y}^{(t+t_{f})}]
\end{equation}
where,
\begin{equation}
\mathbf{y}^{(t)} = [x_{0}^{(t)},y_{0}^{(t)}]
\end{equation}

\noindent are the future co-ordinates of the vehicle being predicted
\subsection{Probabilistic motion prediction}
Our model estimates the conditional distribution $\mbox{P}(\mathbf{Y}|\mathbf{X})$. In order to have the model produce multi-modal distributions, we expand it in terms of maneuvers $m_{i}$, giving:
\begin{equation}
\mbox{P}(\mathbf{Y}|\mathbf{X}) = \sum_{i}\mbox{P}_{\Theta}(\mathbf{Y}|m_{i},\mathbf{X})\mbox{P}(m_{i}|\mathbf{X})
\label{eq:prob}
\end{equation} 
where,
\begin{equation}
\Theta =[\Theta^{(t+1)},..., \Theta^{(t+t_{f})}]
\end{equation}
\noindent are the parameters of a bivariate Gaussian distribution at each time step in the future, corresponding to the means and variances of future locations. 

\subsection{Maneuver classes}
\label{sec:manclasses}
We consider three lateral and two longitudinal maneuver classes as shown in Fig. \ref{fig:form}. The lateral maneuvers consist of left and right lane changes and a lane keeping maneuver. Since lane changes involve preparation and stabilization, we define a vehicle to be in a lane changing state for $\pm$ 4s w.r.t. the actual cross-over.
The longitudinal maneuvers are split into normal driving and braking. We define a vehicle to be performing a braking maneuver if it's average speed over the prediction horizon is less than 0.8 times its speed at the time of prediction. We define our maneuvers in this manner since these maneuver classes are communicated by vehicles to each other through turn signals and brake lights, which will be included as a cue in future work.   

\begin{figure*}[t]
\centering
\includegraphics[width=\textwidth]{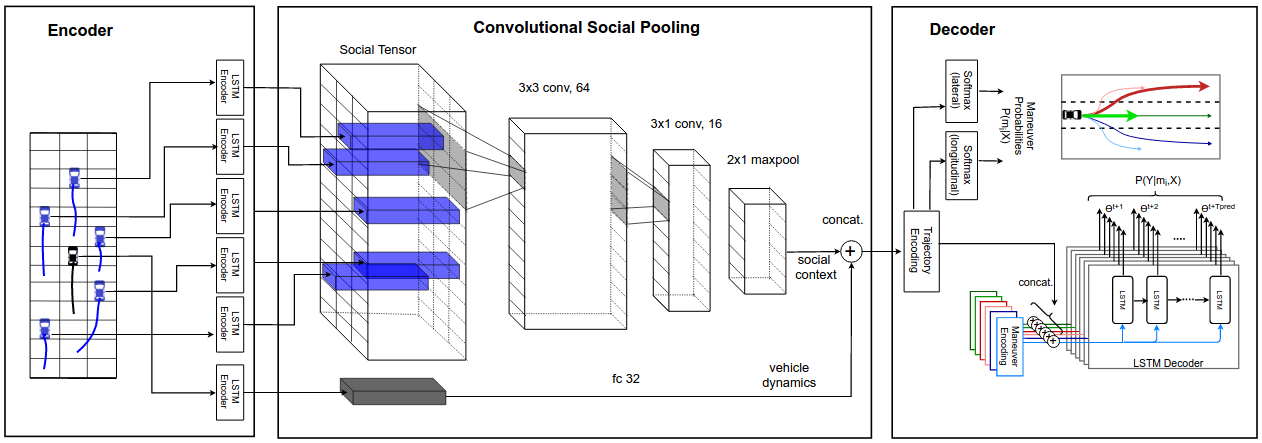}
\caption{\textbf{Proposed Model:} The encoder is an LSTM with shared weights that learns vehicle dynamics based on track histories. The convolutional social pooling layers learn the spatial interdependencies of of the tracks. Finally, the maneuver based decoder outputs a multi-modal predictive distribution for the future motion of the vehicle being predicted}
\label{fig:sys}
\end{figure*}
\section{Model}

Fig. \ref{fig:sys} shows our proposed model. It consists of an LSTM encoder, convolutional social pooling layers and a maneuver based LSTM decoder.

\subsection{LSTM Encoder} 
We use an LSTM encoder for learning the dynamics of vehicle motion. For each instant, snippets of the most recent $t_h$ frames of track history are passed through the LSTM encoder for the vehicle being predicted, and all the vehicles surrounding it. The LSTM states for each vehicle are updated frame by frame over the $t_h$ past frames. The final LSTM state for each vehicle can be expected to encode the state of motion of that vehicle. The LSTMs used for each vehicle have shared weights. This allows for a direct correspondence between the components of the LSTM states for all the vehicles.

\subsection{Convolutional Social Pooling}
\label{sec:csp}
While the LSTM encoder captures the vehicle motion dynamics, it fails to capture the interdependencies of the motion of all vehicles in the scene. Social pooling, proposed in \cite{soclstm}, addresses this by pooling the LSTM states of all the agents around the agent being predicted into a \textit{social tensor}. This is done by defining a spatial grid around the agent being predicted and populating the grid with LSTM states based on the spatial configuration of the agents in the scene. Fig. \ref{fig:sys} shows an example of a social tensor. Using this social tensor as the input to the model in addition to the LSTM state of the agent being predicted, has been shown to improve the accuracy of future motion prediction \cite{soclstm,desire}. This makes sense since the model now gets access to the motion states of surrounding agents and their spatial configuration. 

However, all previous instances of social pooling \cite{soclstm,desire} apply a fully connected layer to the social tensor. This is inefficient since it breaks up the spatial structure of the social tensor. Cells adjacent to each other in space become equivalent to cells far away from each other in the fully connected layer. This can lead to problems in generalization to a test set especially if the agents can be in various different spatial configurations. For example, let's suppose the training set doesn't have a single instance of an LSTM state at spatial location $(m,n)$ of the social tensor. When such an instance is now encountered in the test set, the model will fail to generalize. In particular, this will hold even if there are training instances of LSTM states at spatial grid locations $(m+1,n)$ and $(m,n+1)$, say, in spite of these instances clearly being helpful due to spatial locality. 

As a remedy, we propose the use of convolutional and pooling layers over the social tensor, termed \textit{convolutional social pooling}. The equivariance of the convolutional layers can be expected to help learn locally useful features within the spatial grid of the social tensor, and the max-pooling layer can be expected to add local translational invariance, both of which help address the problem described above. This phenomenon has been further explored in section \ref{sec:exptsec}.

We set up our social tensor by defining a grid based on the lanes. A $13\times3$ spatial grid is defined around the vehicle being predicted, where each column corresponds to a single lane, and the rows are separated by a distance of 15 feet which approximately equals one car length. The social tensor is formed by populating this grid with surrounding car locations. We then apply two convolutional layers and a pooling layer to the social tensor as shown in Fig. \ref{fig:sys} to obtain the social context encoding. Additionally, the LSTM state of the predicted vehicle is passed through a fully connected layer to obtain the vehicle dynamics encoding. The two encodings are concatenated to form the complete trajectory encoding, which is then passed to the decoder.   

\begin{table*}[t]
\centering

\label{tab:results}
\begin{tabular}{@{}ccccccccc@{}}
\toprule
\begin{tabular}[c]{@{}c@{}}Evaluation\\ Metric\end{tabular}         & \begin{tabular}[c]{@{}c@{}}Prediction\\ horizon (s)\end{tabular} & CV   & \begin{tabular}[c]{@{}c@{}}C-VGMM \\+ VIM \cite{self}\end{tabular} & \begin{tabular}[c]{@{}c@{}}GAIL-GRU\\ \cite{gail}\end{tabular} & V-LSTM & \multicolumn{1}{l}{S-LSTM} & \multicolumn{1}{l}{CS-LSTM} & CS-LSTM(M)    \\ \midrule
\multirow{5}{*}{\begin{tabular}[c]{@{}c@{}}RMSE\\ (m)\end{tabular}} & 1                                                                & 0.73 & 0.66                                                        & 0.69                                                    & 0.68   & 0.65                       & \textbf{0.61}               & 0.62          \\
                                                                    & 2                                                                & 1.78 & 1.56                                                        & 1.51                                                    & 1.65   & 1.31                       & \textbf{1.27}               & 1.29          \\
                                                                    & 3                                                                & 3.13 & 2.75                                                        & 2.55                                                    & 2.91   & 2.16                       & \textbf{2.09}               & 2.13          \\
                                                                    & 4                                                                & 4.78 & 4.24                                                        & 3.65                                                    & 4.46   & 3.25                       & \textbf{3.10}               & 3.20          \\
                                                                    & 5                                                                & 6.68 & 5.99                                                        & 4.71                                                    & 6.27   & 4.55                       & \textbf{4.37}               & 4.52          \\ \midrule
\multirow{5}{*}{NLL}                                                & 1                                                                & 3.72 & 2.02                                                        & -                                                       & 1.17   & 1.01                       & 0.89                        & \textbf{0.58} \\
                                                                    & 2                                                                & 5.37 & 3.63                                                        & -                                                       & 2.85   & 2.49                       & 2.43                        & \textbf{2.14} \\
                                                                    & 3                                                                & 6.40 & 4.62                                                        & -                                                       & 3.80   & 3.36                       & 3.30                        & \textbf{3.03} \\
                                                                    & 4                                                                & 7.16 & 5.35                                                        & -                                                       & 4.48   & 4.01                       & 3.97                        & \textbf{3.68} \\
                                                                    & 5                                                                & 7.76 & 5.93                                                        & -                                                       & 4.99   & 4.54                       & 4.51                        & \textbf{4.22} \\ \bottomrule
\end{tabular}
\vspace{0.04 in}
\caption{\textbf{Results:} Root mean squared prediction error (RMSE) and negative log-likelihood (NLL) values over a 5 second prediction horizon for the models being compared. The proposed models outperform the baselines in terms of both RMSE and NLL. Convolutional social pooling leads to lower RMSE values compared to Fully connected social pooling. Using the maneuver based decoder leads to lower NLL values compared to a uni-modal prediction}
\end{table*}
\subsection{Maneuver based LSTM decoder}
We use an LSTM based decoder for generating the predictive distribution for future motion over the next $t_f$ frames. We address the inherent multi-modality of driver behavior by predicting the distribution for each of the six maneuver classes described in section \ref{sec:manclasses} along with the probability for each maneuver class. The decoder has two softmax layers that output the lateral and longitudinal maneuver probabilities. These can be multiplied to give the values of $\mbox{P}(m_{i}|\mathbf{X})$ from Eqn. \ref{eq:prob}. Additionally, an LSTM is used to generate the parameters of a bivariate Gaussian distribution over $t_f$ frames to give the predictive distribution for vehicle motion. In order to obtain maneuver specific distributions $\mbox{P}_{\Theta}(\mathbf{Y}|m_{i},\mathbf{X})$ from Eqn, \ref{eq:prob}, we concatenate the trajectory encoding with a one-hot vector corresponding to the lateral maneuver class and a one-hot vector corresponding to the longitudinal maneuver class.

\subsection{Training and Implementation details}
We train the model end to end. Ideally, we would like to minimize the negative log likelihood \begin{equation}
-\mbox{log}\left(\sum_{i}\mbox{P}_{\Theta}(\mathbf{Y}|m_{i},\mathbf{X})\mbox{P}(m_{i}|\mathbf{X})\right)
\end{equation} 
of the term from from Eqn. \ref{eq:prob} over all the training data points. However, each training instance only provides the realization of one maneuver class that was actually performed. Thus we minimize the negative log likelihood
\begin{equation}
-\mbox{log}\left(\mbox{P}_{\Theta}(\mathbf{Y}|m_{true},\mathbf{X})\mbox{P}(m_{true}|\mathbf{X})\right)
\end{equation}
\noindent over all training instances, instead. 

We train the model using Adam \cite{adam} with learning rate 0.001. The encoder LSTM has 64 dimensional state while the decoder has a 128 dimensional state. The sizes of the convolutional social pooling layers are as shown in Fig. \ref{fig:sys}. The fully connected layer for obtaining the vehicle dynamics encoding has size 32. We use the leaky-ReLU activation with $\alpha$=0.1 for all layers. The model is implemented using PyTorch \cite{pytorch}.

\section{Experimental Evaluation}

\subsection{Dataset}
We use the publicly available NGSIM US-101 \cite{ngsim1} and I-80 \cite{ngsim2} datasets for our experiments. Each dataset consists of trajectories of real freeway traffic captured at 10 Hz over a time span of 45 minutes. Each dataset consists of 15 min segments of mild, moderate and congested traffic conditions. The dataset provides the co-ordinates of vehicles projected to a local co-ordinate system, as defined in section \ref{sec:for}. We split the complete dataset into train and test sets. The test set consists of a fourth of the trajectories from each of the 3 subsets of the US-101 and I-80 datasets. We split the trajectories into segments of 8 s, where we use 3 s of track history and a 5 s prediction horizon. These 8 s segments are sampled at the dataset sampling rate of 10 Hz. However we downsample each segment by a factor of 2 before feeding them to the LSTMs, to reduce the model complexity.

\subsection{Evaluation metrics}

We report results in terms of the root of the mean squared error (RMSE) of the predicted trajectories with respect to the true future trajectories, over a prediction horizon of 5 seconds, as done in \cite{gail}. For the LSTM models generating bivariate Gaussian distributions, the means of the Gaussian components are used for RMSE calculation. For models generating multi-modal predictive distributions, we use the mode with the highest probability for calculating the RMSE. 

While RMSE provides a tangible measure for the predictive accuracy of models, it has limitations while evaluating multi-modal predictions. RMSE is skewed in favor of models that average the modes. In particular, this average may not represent a good prediction. For example, a driver intending to overtake another vehicle may do so by switching to the immediate left or the immediate right lane, while at the same time accelerating. The average of these two modes would be to accelerate while maintaining lane. 

To address this limitation, we additionally report the negative log-likelihood (NLL) of the true trajectories under the predictive distributions generated by the models. While the NLL values cannot be directly interpreted as a physical quantity, they allow us to compare uni-modal and multi-modal predictive distributions.

\subsection{Compared models}
We compare the following baselines and system settings:\\

\noindent \textbf{Baselines:}
\begin{itemize}[itemsep=1pt,parsep=0pt]

\item \textit{Constant Velocity (CV)}: We use a constant velocity Kalman filter as our simplest baseline

\item \textit{C-VGMM + VIM:} We use maneuver based variational Gaussian mixture models with a Markov random field based vehicle interaction module described in \cite{self} as our second baseline. We modify the model to use the maneuver classes described in this work to allow for a fair comparison

\item \textit{GAIL-GRU:} We consider the generative adversarial imitation learning model described in \cite{gail}. Since the same datasets have been used in both works, we use the results reported by the authors in the original article. There is a caveat that the GAIL-GRU trajectories were generated by running the policy one vehicle at a time, while all surrounding vehicles move according to the ground-truth of the NGSIM dataset. Thus, the model has access to the true trajectories of adjacent vehicles over the prediction horizon.
\end{itemize}

\noindent\textbf{System settings:}
\begin{itemize}[itemsep=1pt,parsep=0pt]
\item \textit{Vanilla LSTM (V-LSTM)}: This simply uses the track history of the predicted vehicle in the encoder LSTM and generates a unimodal output distribution with the LSTM decoder
\item \textit{LSTM with fully connected social pooling (S-LSTM)}: This uses the fully connected social pooling described in \cite{soclstm} and generates a unimodal output distribution
\item \textit{LSTM with convolutional social pooling (CS-LSTM)}: This uses convolutional social pooling and generates a unimodal output distribution
\item \textit{LSTM with convolutional social pooling and maneuvers (CS-LSTM(M))}: This is the complete model described in this paper, including the maneuver based decoder generating a multi-modal predictive distribution  
 
\end{itemize}

\subsection{Results}
\label{sec:res}

Table 1 shows the RMSE and NLL values for the models being compared. S-LSTM, CS-LSTM, and CS-LSTM(M) outperform the baselines \cite{self,gail} in terms of RMSE and NLL values, showing the effectiveness of the proposed model. 

We note that the vanilla LSTM and CV models produce higher RMSE values compared to the other models. Each of the other models use some information about the motion of neighboring vehicles. This shows that inter-vehicle interaction is a useful cue for motion prediction, consistent with the results reported in \cite{soclstm, desire, self}.

We also note that CS-LSTM outperforms the S-LSTM in terms of both RMSE and NLL values. This suggests that convolutional social pooling better models the interdependencies of vehicle motion compared to a fully connected social pooling layer. We further analyze this in the following section. 

Finally, we note that CS-LSTM(M) leads to higher RMSE values compared to CS-LSTM. This could, in part, be due to misclassified maneuvers, since the RMSE values for CS-LSTM(M) are calculated using the trajectory corresponding to the maneuver with the highest probability. However we note that CS-LSTM(M) achieves significantly lower NLL values compared to CS-LSTM. Thus the predictive distribution generated by CS-LSTM(M) better fits the true trajectories compared to that generated by CS-LSTM. This points to the multi-modal nature of the task.

\subsection{Fully connected vs. convolutional social pooling}
\begin{figure}[t]
\centering
\includegraphics[width=\columnwidth]{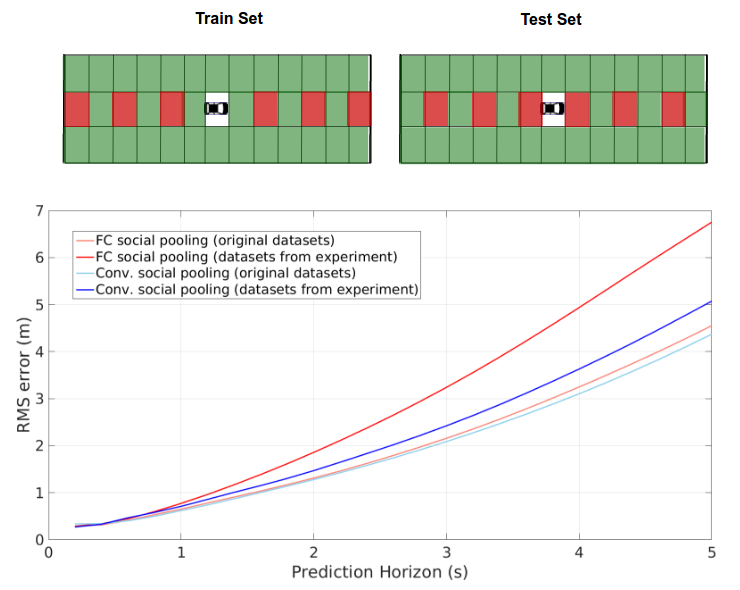}
\caption{Experiment comparing fully connected and convolutional social pooling. \textbf{Top:} All training instances with vehicles at odd locations in ego lane of social tensor removed from train set; all instances with vehicles even locations removed from test set. \textbf{Bottom:} RMS values of prediction error for FC social pooling and convolutional social pooling for original datasets and datasets from experiment. Convolutional social pooling is more robust to missing spatial patterns in the social tensor}
\label{fig:expt}
\end{figure}

\label{sec:exptsec}
We conjectured in section \ref{sec:csp} that fully connected social pooling as described in \cite{soclstm} would poorly generalize to a test set with even slight differences in spatial patterns of agents in the scene as collected in the social tensor, and that convolutional social pooling would remedy this. The reduced prediction error from section \ref{sec:res} seems to suggest that this is true. However to further analyze this, we set up the following experiment. We remove all instances from the train set corresponding to the odd grid locations of vehicles from the ego lane, and remove all instances from the test set corresponding to even grid locations as shown in Fig. \ref{fig:expt}. Thus, we have a train and test set with zero overlap in terms of spatial configurations of the social tensors. However, we have plenty of spatially similar but not identical configurations common to both. We plot the RMS values of prediction error for this new train and test set, for fully connected social pooling and convolutional social pooling models. We see that the performance of the fully connected social pooling model drastically drops, almost to the point of the vanilla LSTM shown in section \ref{sec:res}. The performance drop with convolutional social pooling is less severe in comparison. This suggests that using convolutional and pooling layers to aggregate social context is a much more robust approach compared to using a fully connected layer.

\subsection{Qualitative analysis of predictions}
\begin{figure*}[t]
\centering
\includegraphics[width=\textwidth]{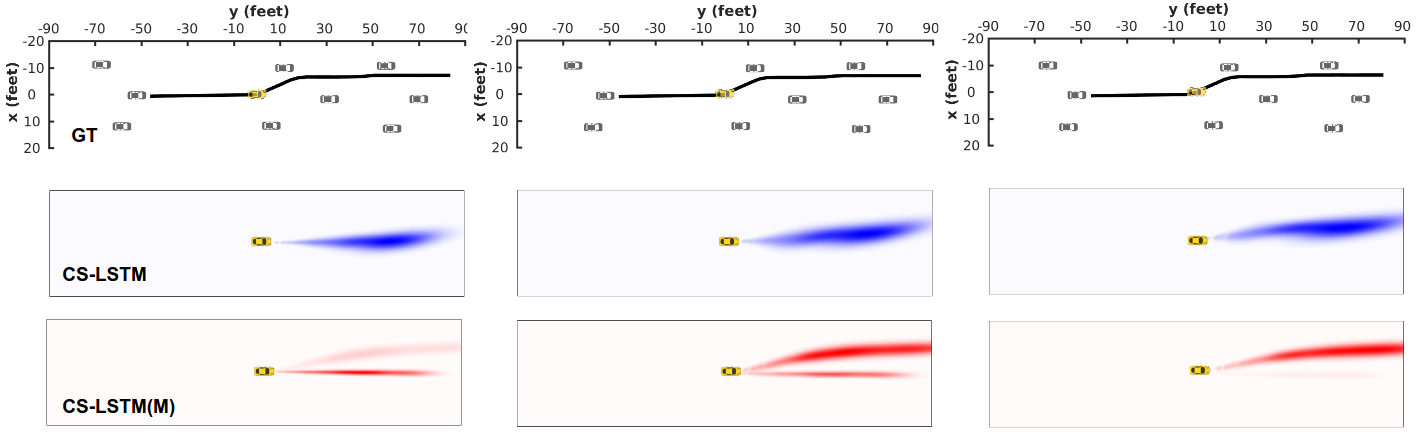}
\caption{ \textbf{Comparison of uni-modal and multi-modal predictions:} The figure shows the true trajectory (top, black), CS-LSTM predictive distributions (middle, blue) and CS-LSTM(M) predictive distributions (bottom, red) for three consecutive frames of a lane change maneuver. The heat maps are generated by plotting the Gaussian components for each maneuver at each time step in the prediction horizon}
\label{fig:ex1}
\end{figure*}

In this section we qualitatively analyze the predictions made by our model to gain insights into its behavior in various traffic configurations.\\ 

\noindent\textbf{Uni-modal vs. multi-modal predictions:} Figure \ref{fig:ex1} shows a comparison of the unimodal predictive distribution generated by CS-LSTM and the multi-modal distribution generated by CS-LSTM(M). The plots show three consecutive frames during a lane change maneuver from left to right. The top row shows the track history and the true future trajectory. The middle row shows the predictive distribution generated by CS-LSTM and the bottom row shows the predictive distribution generated by CS-LSTM(M). We can clearly observe two modes in the predictive distribution of CS-LSTM(M). The mode corresponding to the lane change becomes more and more prominent further into the maneuver while the mode corresponding to the keep lane maneuver fades away. We further note that for all three cases, the mode corresponding to the lane change closely matches the true future trajectory. However, the unimodal distribution generated by CS-LSTM shows an average of the two modes and also has greater variance. This illustrates why the CS-LSTM achieves lower RMSE values while leading to higher NLL values as compared to CS-LSTM(M). \\

\begin{figure*}[t]
\centering
\includegraphics[width=\textwidth]{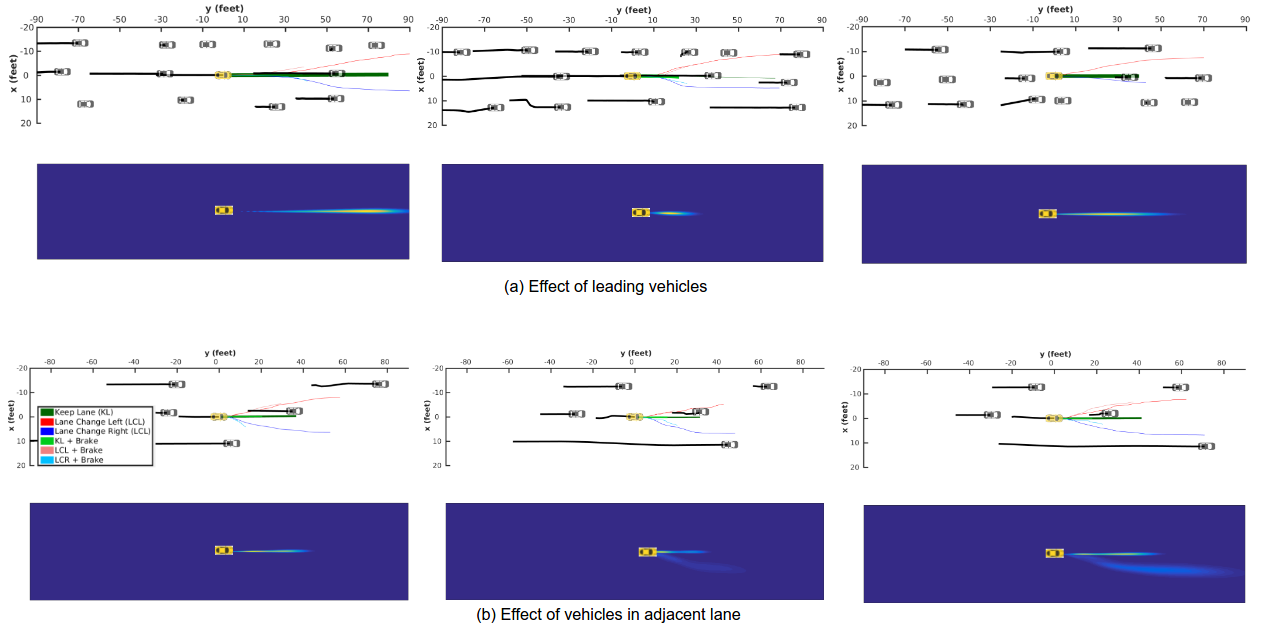}
\caption{ \textbf{Surrounding vehicles affect predictions:} This figure shows the effect of surrounding vehicles on predictive distribution generated by the model. The heat maps are generated by plotting the Gaussian components for each maneuver at each time step in the prediction horizon}
\label{fig:ex2}
\end{figure*}

\noindent \textbf{Effect of surrounding vehicles on predictiions:} Figure \ref{fig:ex2} shows six different scenarios of traffic. Each figure shows a plot of track histories over the past 3 seconds and the mean predicted trajectories over the next 5 seconds for each maneuver class. The thickness of the plots of the predicted trajectories is proportional to the probabilities assigned to each maneuver class. Additionally, each figure shows a heat map of the complete predicted distribution.

Fig 6(a) shows the effect of the leading vehicle on the predictions made by the model. The first example (top-left) shows an example of free flowing traffic, where the predicted vehicle and the leading vehicle are moving at approximately the same speed. In the second example (top-middle), we note from the track histories that the leading vehicles are slowing down compared to the predicted vehicle. We see that the model predicts the vehicle to brake, although it's current motion suggests otherwise. Conversely, in the third example (top-right), we see that the vehicle being predicted is almost stationary, while the leading vehicles are beginning to move. The model predicts the vehicle to accelerate, as is expected in stop-and-go traffic.

Fig 6(b) shows the effect of vehicles in the adjacent lane on the model's predictions. The three examples show the same scenario separated by 0.5 s. We note that the vehicle being predicted is in a congested lane, with its leading vehicle slowing down. We also note that the adjacent left lane is congested. On the other hand, the adjacent right lane is moving at a much faster speed. Based on this, the model assigns a high probability to the predicted vehicle staying in lane and braking, as expected. However, it also assigns a small probability to an overtake by moving to the right lane. We can observe that the model assigns a greater probability to the overtake as the adjacent vehicle moves further away, clearing up the lane. 

\section{Conclusions and Future Work}
We have presented an LSTM encoder-decoder based model for vehicle trajectory prediction for reasoning about the interdependencies neighboring vehicles' motion. Our model uses an improved social pooling layer using convolutional connections as opposed to fully connected layers that more robustly models and better generalizes the various spatial configurations of interacting agents in a scene. We term this \textit{convolutional social pooling}. Our proposed model outperforms the reported state of the art on two large publicly available datasets of vehicle trajectories. Our model outputs multi-modal distributions for future motion of vehicles based on maneuver classes. We have presented a qualitative analysis of the predicted distributions.  

One limitation of the current approach is that it relies purely on vehicle tracks to infer maneuver classes and future trajectories. A lot of complementary information can be captured using visual and map based cues. These could be used to improve the accuracy of maneuver classification and thus, that of future motion prediction. Future work would focus on incorporation of these cues into the model

\section{Acknowledgement}
We would like to thank Ishan Gupta and the anonymous reviewers for their useful inputs. We also gratefully acknowledge the continued support of our industry sponsors.

{\small
\bibliographystyle{ieee}
\bibliography{egbib}

\begin{thebibliography}{10}\itemsep=-1pt

\bibitem{soclstm}
A.~Alahi, K.~Goel, V.~Ramanathan, A.~Robicquet, L.~Fei-Fei, and S.~Savarese.
\newblock Social lstm: Human trajectory prediction in crowded spaces.
\newblock In {\em Proceedings of the IEEE Conference on Computer Vision and
  Pattern Recognition}, pages 961--971, 2016.

\bibitem{bahram}
M.~Bahram, C.~Hubmann, A.~Lawitzky, M.~Aeberhard, and D.~Wollherr.
\newblock A combined model-and learning-based framework for interaction-aware
  maneuver prediction.
\newblock {\em IEEE Transactions on Intelligent Transportation Systems},
  17(6):1538--1550, 2016.

\bibitem{cho}
K.~Cho, B.~Van~Merri{\"e}nboer, C.~Gulcehre, D.~Bahdanau, F.~Bougares,
  H.~Schwenk, and Y.~Bengio.
\newblock Learning phrase representations using rnn encoder-decoder for
  statistical machine translation.
\newblock {\em arXiv preprint arXiv:1406.1078}, 2014.

\bibitem{ngsim1}
J.~Colyar and J.~Halkias.
\newblock Us highway 101 dataset.
\newblock {\em Federal Highway Administration (FHWA), Tech. Rep.
  FHWA-HRT-07-030}, 2007.

\bibitem{ngsim2}
J.~Colyar and J.~Halkias.
\newblock Us highway i-80 dataset.
\newblock {\em Federal Highway Administration (FHWA), Tech. Rep.
  FHWA-HRT-07-030}, 2007.

\bibitem{self}
N.~Deo, A.~Rangesh, and M.~M. Trivedi.
\newblock How would surround vehicles move? a unified framework for maneuver
  classification and motion prediction.
\newblock {\em arXiv preprint arXiv:1801.06523}, 2018.

\bibitem{graves}
A.~Graves.
\newblock Generating sequences with recurrent neural networks.
\newblock {\em arXiv preprint arXiv:1308.0850}, 2013.

\bibitem{houenou}
A.~Houenou, P.~Bonnifait, V.~Cherfaoui, and W.~Yao.
\newblock Vehicle trajectory prediction based on motion model and maneuver
  recognition.
\newblock In {\em Intelligent Robots and Systems (IROS), 2013 IEEE/RSJ
  International Conference on}, pages 4363--4369. IEEE, 2013.

\bibitem{vi1}
E.~K{\"a}fer, C.~Hermes, C.~W{\"o}hler, H.~Ritter, and F.~Kummert.
\newblock Recognition of situation classes at road intersections.
\newblock In {\em Robotics and Automation (ICRA), 2010 IEEE International
  Conference on}, pages 3960--3965. IEEE, 2010.

\bibitem{aida}
A.~Khosroshahi, E.~Ohn-Bar, and M.~M. Trivedi.
\newblock Surround vehicles trajectory analysis with recurrent neural networks.
\newblock In {\em Intelligent Transportation Systems (ITSC), 2016 IEEE 19th
  International Conference on}, pages 2267--2272. IEEE, 2016.

\bibitem{kim}
B.~Kim, C.~M. Kang, S.~H. Lee, H.~Chae, J.~Kim, C.~C. Chung, and J.~W. Choi.
\newblock Probabilistic vehicle trajectory prediction over occupancy grid map
  via recurrent neural network.
\newblock {\em arXiv preprint arXiv:1704.07049}, 2017.

\bibitem{adam}
D.~P. Kingma and J.~Ba.
\newblock Adam: A method for stochastic optimization.
\newblock {\em arXiv preprint arXiv:1412.6980}, 2014.

\bibitem{gail}
A.~Kuefler, J.~Morton, T.~Wheeler, and M.~Kochenderfer.
\newblock Imitating driver behavior with generative adversarial networks.
\newblock In {\em Intelligent Vehicles Symposium (IV), 2017 IEEE}, pages
  204--211. IEEE, 2017.

\bibitem{laug}
C.~Laugier, I.~E. Paromtchik, M.~Perrollaz, M.~Yong, J.-D. Yoder, C.~Tay,
  K.~Mekhnacha, and A.~N{\`e}gre.
\newblock Probabilistic analysis of dynamic scenes and collision risks
  assessment to improve driving safety.
\newblock {\em IEEE Intelligent Transportation Systems Magazine}, 3(4):4--19,
  2011.

\bibitem{desire}
N.~Lee, W.~Choi, P.~Vernaza, C.~B. Choy, P.~H. Torr, and M.~Chandraker.
\newblock Desire: Distant future prediction in dynamic scenes with interacting
  agents.
\newblock 2017.

\bibitem{surv}
S.~Lef{\`e}vre, D.~Vasquez, and C.~Laugier.
\newblock A survey on motion prediction and risk assessment for intelligent
  vehicles.
\newblock {\em Robomech Journal}, 1(1):1, 2014.

\bibitem{nilsson}
J.~Nilsson, J.~Silvlin, M.~Brannstrom, E.~Coelingh, and J.~Fredriksson.
\newblock If, when, and how to perform lane change maneuvers on highways.
\newblock {\em IEEE Intelligent Transportation Systems Magazine}, 8(4):68--78,
  2016.

\bibitem{pytorch}
A.~Paszke, S.~Gross, S.~Chintala, G.~Chanan, E.~Yang, Z.~DeVito, Z.~Lin,
  A.~Desmaison, L.~Antiga, and A.~Lerer.
\newblock Automatic differentiation in pytorch.
\newblock 2017.

\bibitem{phil}
D.~J. Phillips, T.~A. Wheeler, and M.~J. Kochenderfer.
\newblock Generalizable intention prediction of human drivers at intersections.
\newblock In {\em Intelligent Vehicles Symposium (IV), 2017 IEEE}, pages
  1665--1670. IEEE, 2017.

\bibitem{gmrlat}
J.~Schlechtriemen, F.~Wirthmueller, A.~Wedel, G.~Breuel, and K.-D. Kuhnert.
\newblock When will it change the lane? a probabilistic regression approach for
  rarely occurring events.
\newblock In {\em Intelligent Vehicles Symposium (IV), 2015 IEEE}, pages
  1373--1379. IEEE, 2015.

\bibitem{schreier}
M.~Schreier, V.~Willert, and J.~Adamy.
\newblock Bayesian, maneuver-based, long-term trajectory prediction and
  criticality assessment for driver assistance systems.
\newblock In {\em Intelligent Transportation Systems (ITSC), 2014 IEEE 17th
  International Conference on}, pages 334--341. IEEE, 2014.

\bibitem{dpdm}
S.~Sivaraman and M.~M. Trivedi.
\newblock Dynamic probabilistic drivability maps for lane change and merge
  driver assistance.
\newblock {\em IEEE Transactions on Intelligent Transportation Systems},
  15(5):2063--2073, 2014.

\bibitem{tran}
Q.~Tran and J.~Firl.
\newblock Online maneuver recognition and multimodal trajectory prediction for
  intersection assistance using non-parametric regression.
\newblock In {\em Intelligent Vehicles Symposium Proceedings, 2014 IEEE}, pages
  918--923. IEEE, 2014.

\bibitem{ulbrich}
S.~Ulbrich and M.~Maurer.
\newblock Towards tactical lane change behavior planning for automated
  vehicles.
\newblock In {\em Intelligent Transportation Systems (ITSC), 2015 IEEE 18th
  International Conference on}, pages 989--995. IEEE, 2015.

\end{thebibliography}
}

\end{document}